\documentclass{article}
\usepackage{psfig}

\begin{document}

\title{A proposal to design expert system for the calculations in the domain of QFT}%
\author{Andrea Severe S. \footnote{sys01@narod.ru; Russia, Protvino}} %

\maketitle
\begin{abstract}
Main purposes of the paper are followings: 1) To show examples of
the calculations in domain of QFT via ``derivative rules'' of an
expert system; 2) To consider advantages and disadvantage that
technology of the calculations; 3) To reflect about how one would
develop new physical theories, what knowledge would be useful in
their investigations and how this problem can be connected with
designing an expert system.
\end{abstract}

\section{Introduction}
An analytical-calculation technology is a manipulation technique
of the symbolic expressions, corresponding some mathematical
objects, by digital computers \cite{CAS}. As examples of the
realization of the symbolic calculations systems one can name most
popular systems such as: MAXIMA (http://maxima.sourceforge.net/,
"5.9.0", 2003), { \sl Mathematica } (http://www.wolfram.com, "5",
2003), Maple (http://www.maplesoft.com , "9.0", 2003), muPAD
(http://www.mupad.de/index\_uni.shtml, "2.5", 2002). General ideas
engrained in these systems are usually referred as Computer
Algebra technique (CA) and systems themselves are called CASs.

CAS usually provides general environment for manipulation of the
symbolic expressions, and for the calculations in specific
knowledge domain special-purpose packages are created. For
example, FeynCalc package \cite{FeynC} for {\sl Mathematica}
system provides facilities to do calculation upon objects in QFT.

The packages extending field of application of the CAS are usually
worked out in procedural-programming paradigm. That is each
package usually provides a number of functions which can be used.
However it frequently happens that user should transform an answer
which the employed function give out, and transformation chain can
be nontrivial \cite{Wester}. Moreover let's noted that a number of
the possible transformation functions can be large.

By using expert system's \cite{ES-ref} ``derivation rules''
technique one can find desirable answer by calling only one
function - a searching function (and a searching time certainly
increases to one or more powers in the value in return). Besides
the simple convenience this possibility opens new outlook: since
no human assistance is needed for the program to find an answer
then one could pose a problem of next generation of the
complexity. For example, it will be possible to design the systems
which could formulate a searching task or could determine
equivalence of several searching tasks. ``Consideration'' section
of the paper is devoted to this problem.

There are common prejudices about expert systems and their
application in the scientific research. ``Appendix A'' contains
two very simple examples of the calculation via ``derivation
rules''.

\section{Consideration$^1$}
\footnotetext[1]{None of the scientific approaches were used in
this section. Nevertheless the author thinks that the
considerations could be interesting.}

One can propose the following structure of the physical knowledge:
$$ <Phys.Problem\{i\}>+<Phys.Theory>\Longrightarrow q_{i}, $$
where $<Phys.Problem\{i\}>$ -- is a formulation of the physical
problem,\\
 {`$+<Phys.Theory>$'} means --- in term of the certain
physical theory; $q_{i}$ -- are the mathematical expressions that
can be probed experimentally.

In order to do computer calculations symbolic representation of
the physical and mathematical objects should be given:
$$<Phys.Problem\{i\}>\longrightarrow \rm [Problem\{i\}]$$
$$<Phys.Theory>\longrightarrow \rm [ ES ],$$
$$q_{i}\longrightarrow \rm [q_{i}],$$ where [...] assigns symbolic
representation of the object `...'. Let's imagine that
$<Phys.Theory>$ can be represented as an expert system. Then
calculation can be expressed as: $$\rm [ q_{i}==ES( Problem\{i\}
)]$$

These expressions can be understood as current state of out
knowledge expressed in term of some computer system (for example,
CAS). Any increase or modification of the physical knowledge
should be reflected in form of that expressions (opposite
statement is not take place).

Let's assume that some computer symbolic transformation can be
constructed such as: $$\rm [ ES \rightarrow ES'],$$
$$\rm[Problem\{i\}\rightarrow Problem\{i\}'],$$ $$\rm
[q_i\rightarrow q_i + \Delta_i(ES,Task) ],$$ where numeric values
of the $\Delta_i$ less then uncertainty of the $<q_i>$. And
moreover: $${\rm [ES'] \longrightarrow} <Phys.Theory'> $$ then one
could conclude that $<Phys.Theory'>$ can be treated as new
physical theory.

In the end of the section author would like to notice a reason why
QFT is especial interesting domain for designing the expert
system. If ES for the calculation in QFT would be created one
could ask a question why only this semantic structure
corresponding to the structure of the ES's symbolic productions is
marked out in nature. May be realization of the reason will be
helpful in better understanding of nature. The more fundamental
theory is taken to build an expert system for the more fundamental
semantic structure of the knowledge of the ES can be investigated.

\section{Conclusions}

This paper eventually contains more questions then answers. Sorry
for that. Nevertheless let's do some summary in following items:

\begin{itemize}
\item Examples of the calculations in QFT domain via ``derivative rules'' of an
expert system are presented in Appendices A and B;

\item  Advantages of that technology of the calculations are as
followings: full automation of the calculation in some specific
domain (and since increase labour productivity of the theorists);
new programming architecture which will demand further
investigations; new outlook in the investigation of the physical
theories.

\item Disadvantages of that technology of the calculations are as
followings: huge increase of the time of the calculations; large
cost in term of man-year-intelligence.

\end{itemize}

\section{Acknowledgements}

I would like to thank my mother and friends.

\noindent This project {\bf{was not}} supported by RFBF by the
project 04-07-90165.

\section{Appendix A}
This appendix contains very simple examples of the calculation via ``derivation rules''
and  exhibits  calculation of the Momentum-Energy tensor in the theory of free
electro-magnetic field. First we calculate non-symmetry TEM ($T[\mu,\nu]$), check it's
property and then symmetrize it ($T1[\mu,\nu]$). This standard academic task is taken
from \cite{Peskin}.

Searching procedure has name `Search'.

The examples can be downloaded from http://sys01.narod.ru/fc-1-10.nb

\psfig{file=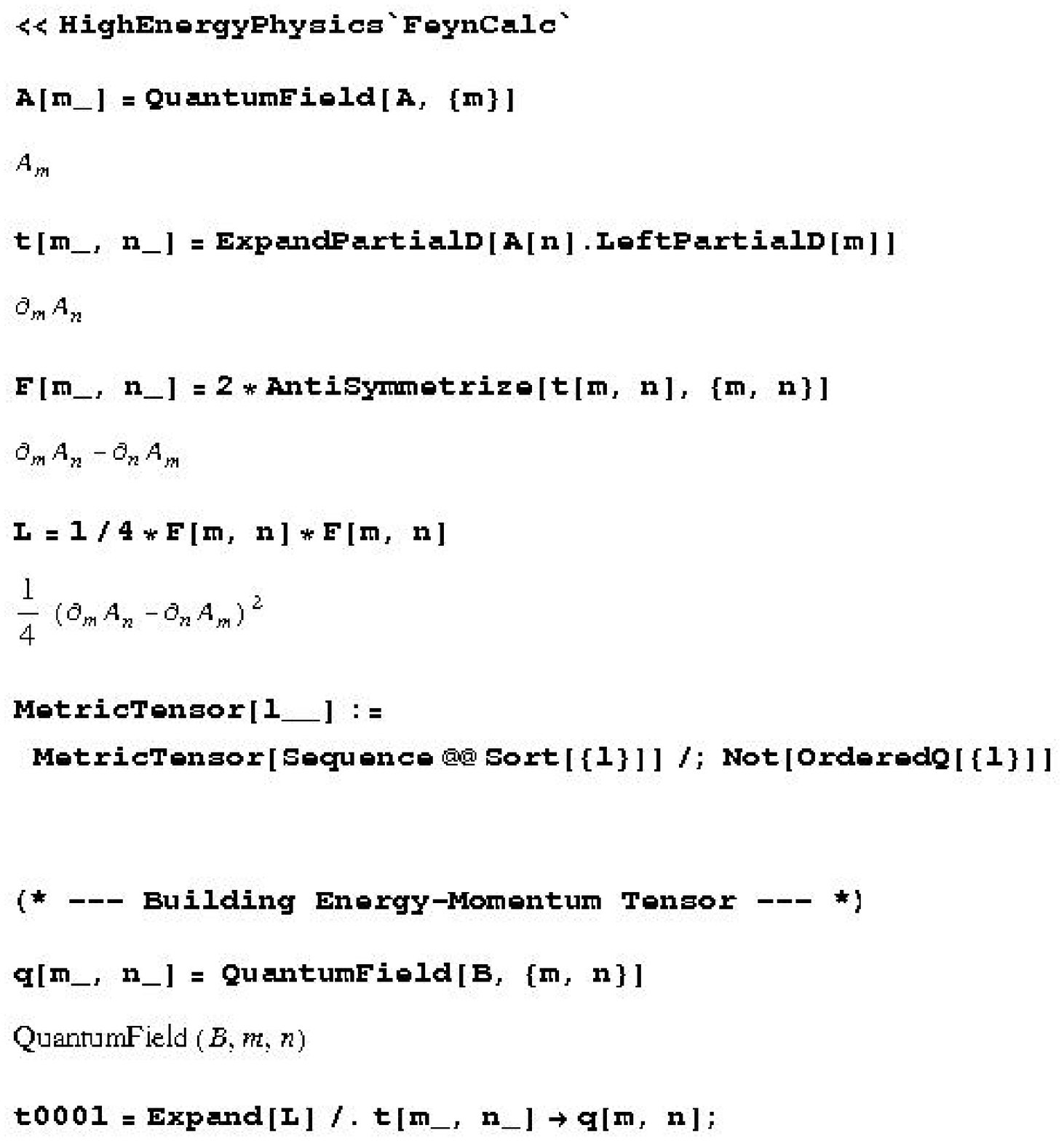,width=300pt}

\psfig{file=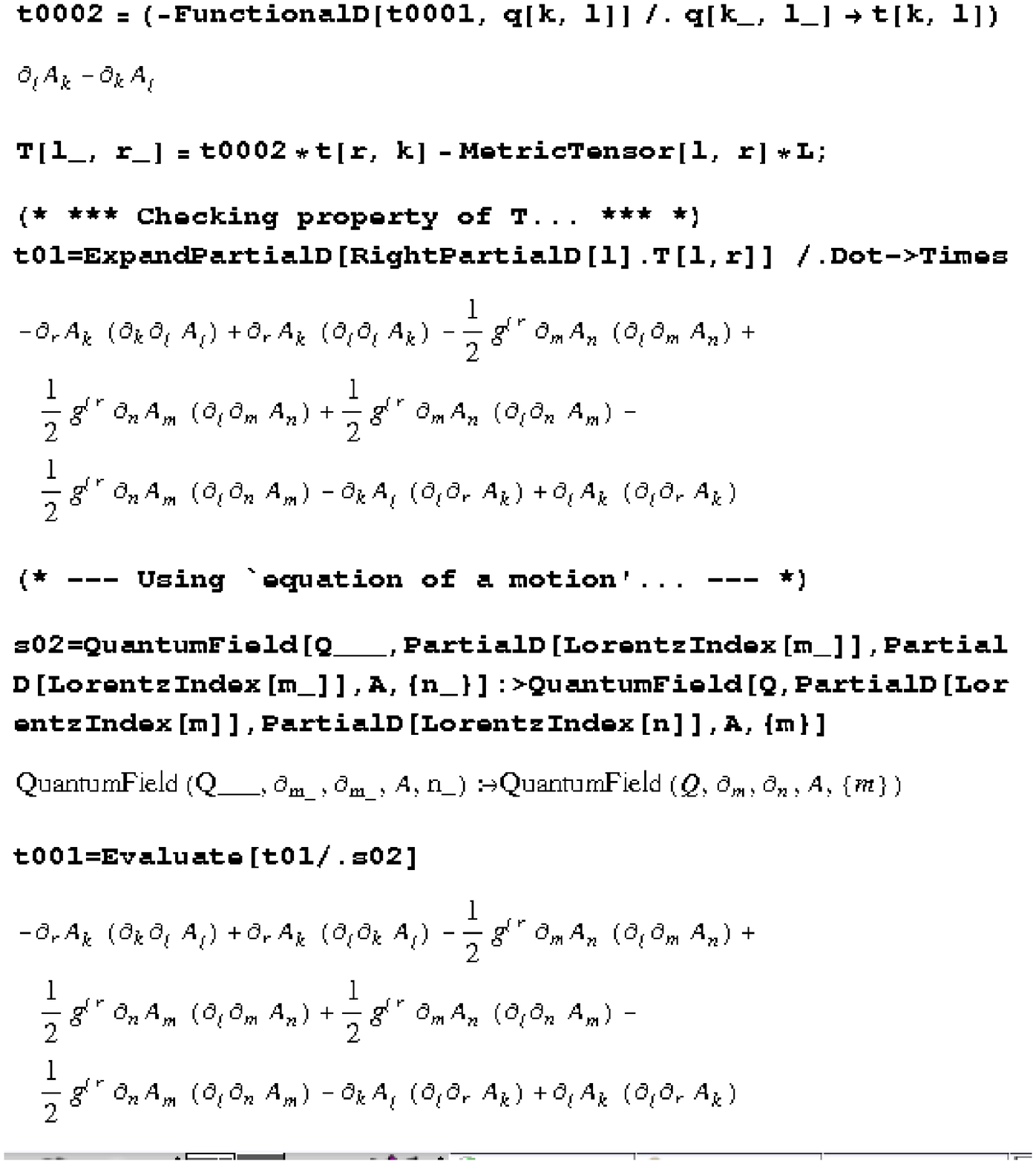,width=300pt}

\psfig{file=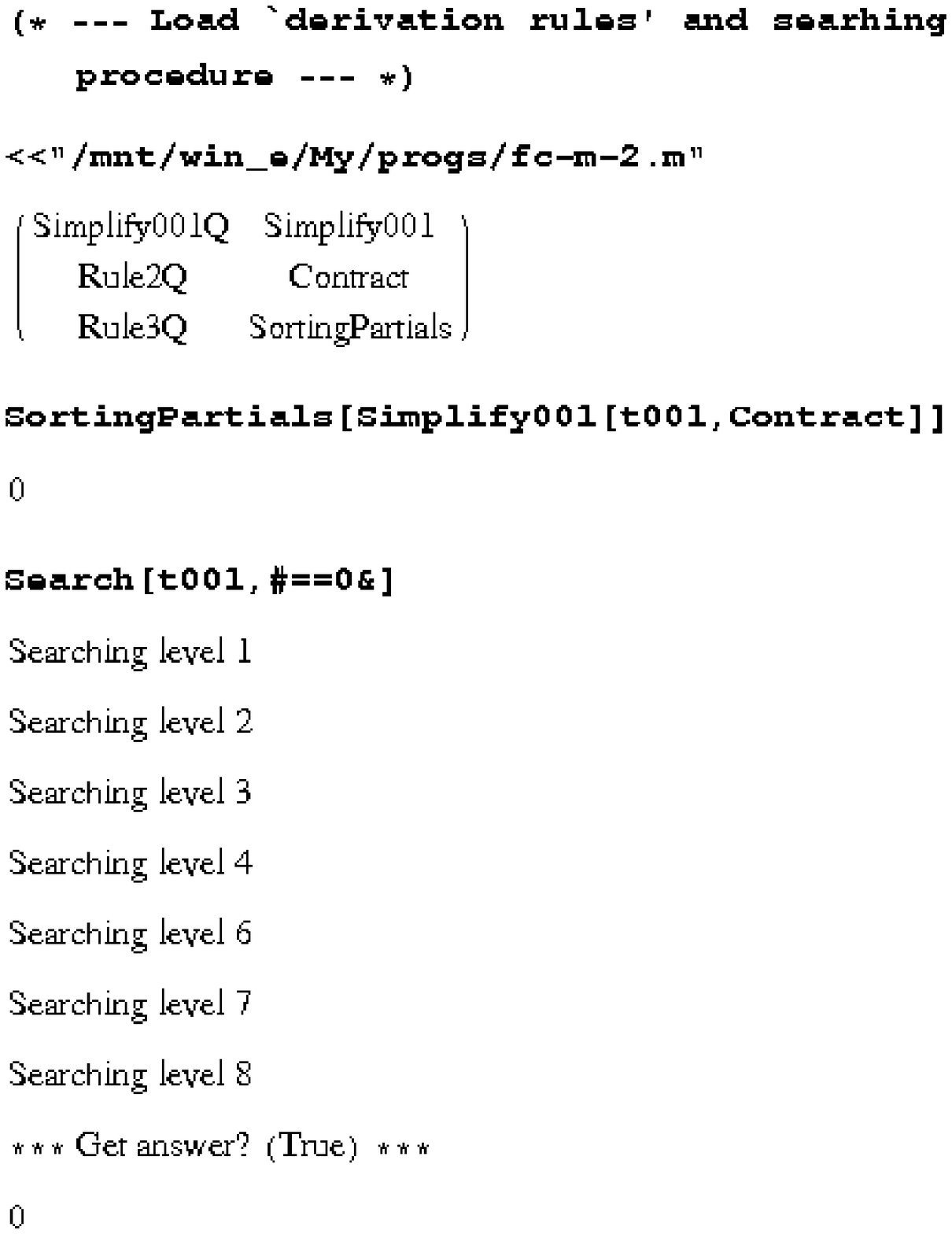,width=300pt}

\psfig{file=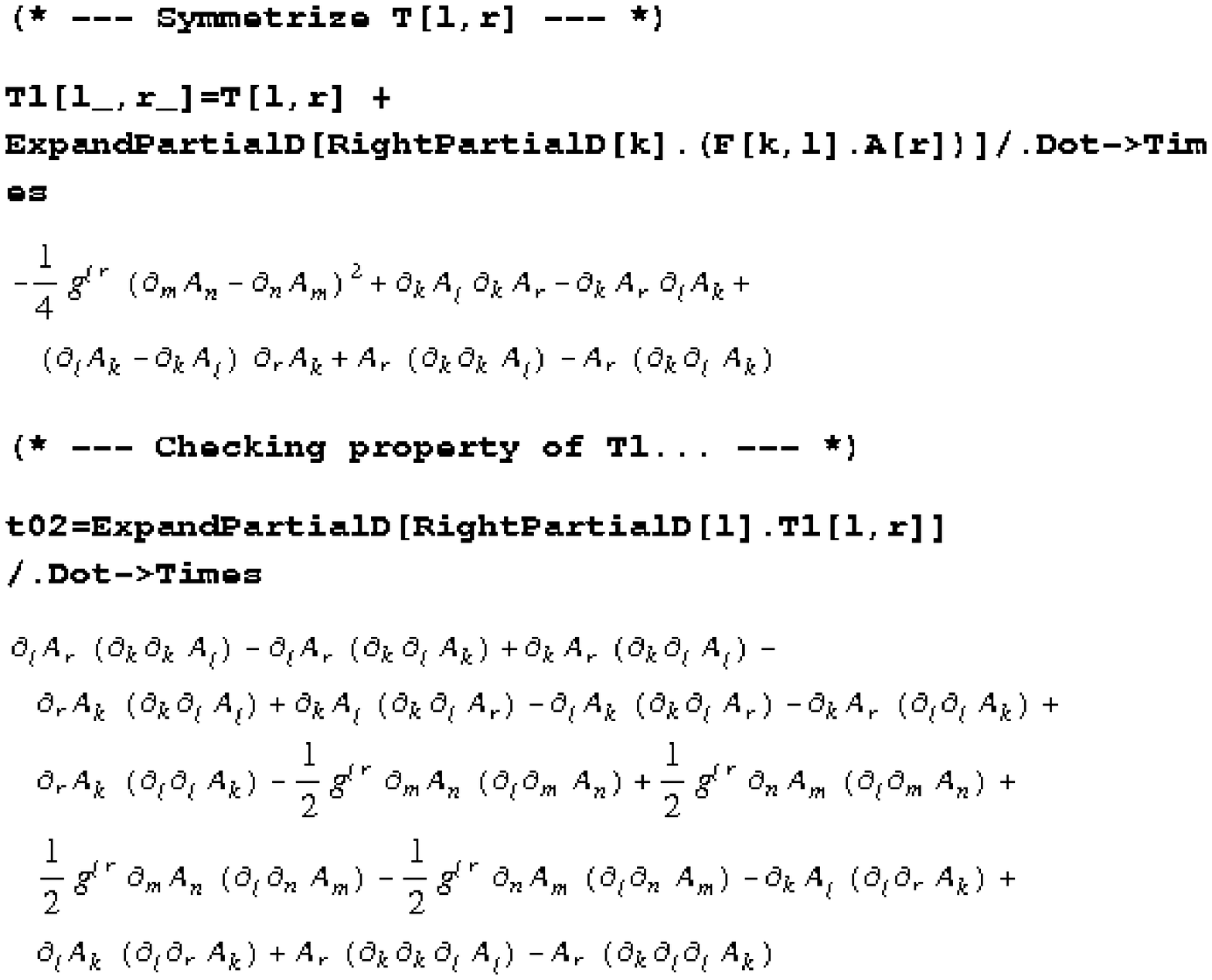,width=300pt}

\psfig{file=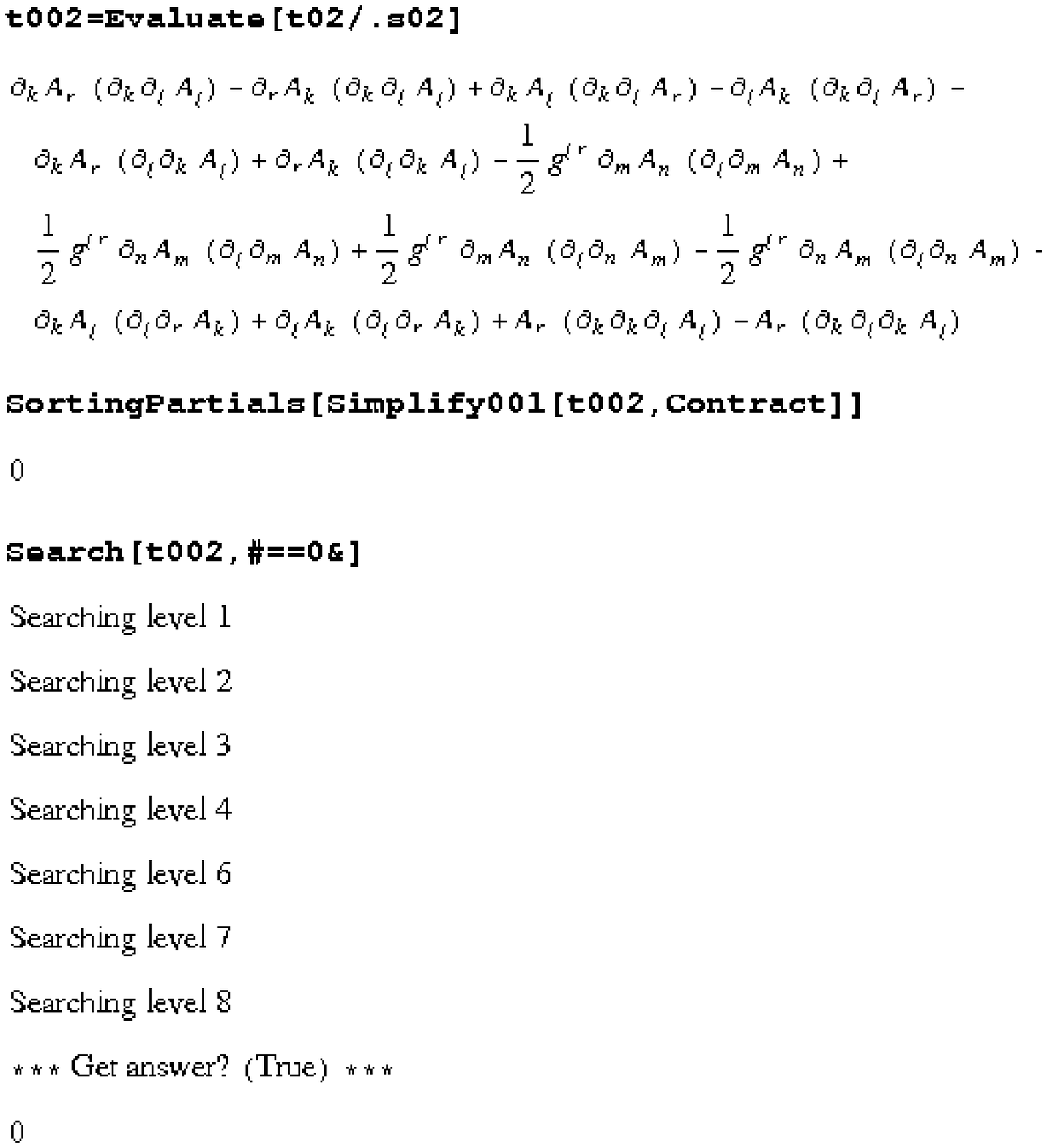,width=300pt}

\section{Appendix B}
This appendix should contain the module which is used in examples from Appendix A. It's
available at http://sys01.narod.ru/fc-m-2.m

Searching procedure has name `Search'. Breadth first search algorithm was used combined
with cycling control.

Only three ``derivation rules'' were created and stored in the variable `Rules'. User's
rules can be added to it simply.

\end{document}